\documentclass[10pt,twocolumn,letterpaper]{article}

\usepackage{cvpr}
\usepackage{times}
\usepackage{epsfig}
\usepackage{graphicx}
\usepackage{amsmath}
\usepackage{amssymb}
\usepackage{bbding}
\usepackage{paralist}
\usepackage[pagebackref=true,breaklinks=true,letterpaper=true,colorlinks,citecolor=blue,linkcolor=blue,bookmarks=false]{hyperref}
\usepackage{bbding}
\usepackage{textcomp,booktabs}
\usepackage[usenames,dvipsnames]{color}
\usepackage{colortbl}
\definecolor{mygray}{gray}{.9}
\definecolor{mypink}{rgb}{.99,.91,.95}
\definecolor{mycyan}{cmyk}{.3,0,0,0}
\usepackage{multirow}

\cvprfinalcopy 

\def\cvprPaperID{423} 

\begin{document}
	
	\title{Fast and Accurate Online Video Object Segmentation via Tracking Parts}
    
  \author{Jingchun Cheng$^{1,2}$
    \hspace{0.1in} Yi-Hsuan Tsai$^{3}$ 
    \hspace{0.1in} Wei-Chih Hung$^{2}$
	\hspace{0.1in} Shengjin Wang$^{1}$\thanks{Corresponding Author}
	\hspace{0.1in} Ming-Hsuan Yang$^{2}$
	\vspace{1mm} \\
	\hspace{0.1in} $^{1}$Tsinghua University \hspace{0.10in} $^{2}$University of California, Merced
		\hspace{0.1in} $^{3}$NEC Laboratories America\\
      }
	
	\maketitle
	
	\begin{abstract}
		Online video object segmentation is a challenging task as it entails to process the image sequence timely and accurately.
		To segment a target object through the video, numerous CNN-based methods have been developed by heavily finetuning on the object mask in the first frame, which is time-consuming for online applications.
		In this paper, we propose a fast and accurate video object segmentation algorithm that can immediately start the segmentation process once receiving the images.
		We first utilize a part-based tracking method to deal with challenging factors such as large deformation, occlusion, and cluttered background.
		Based on the tracked bounding boxes of parts, we construct a region-of-interest segmentation network to generate part masks.
		Finally, a similarity-based scoring function is adopted to refine these object parts by comparing them to the 
		visual information in the first frame.
		Our method performs favorably against state-of-the-art algorithms in accuracy on the DAVIS benchmark dataset, while achieving much faster runtime performance.
	\end{abstract}
	
	\section{Introduction}
\begin{figure}[t]
		\begin{center}
			\includegraphics[width=1\linewidth]{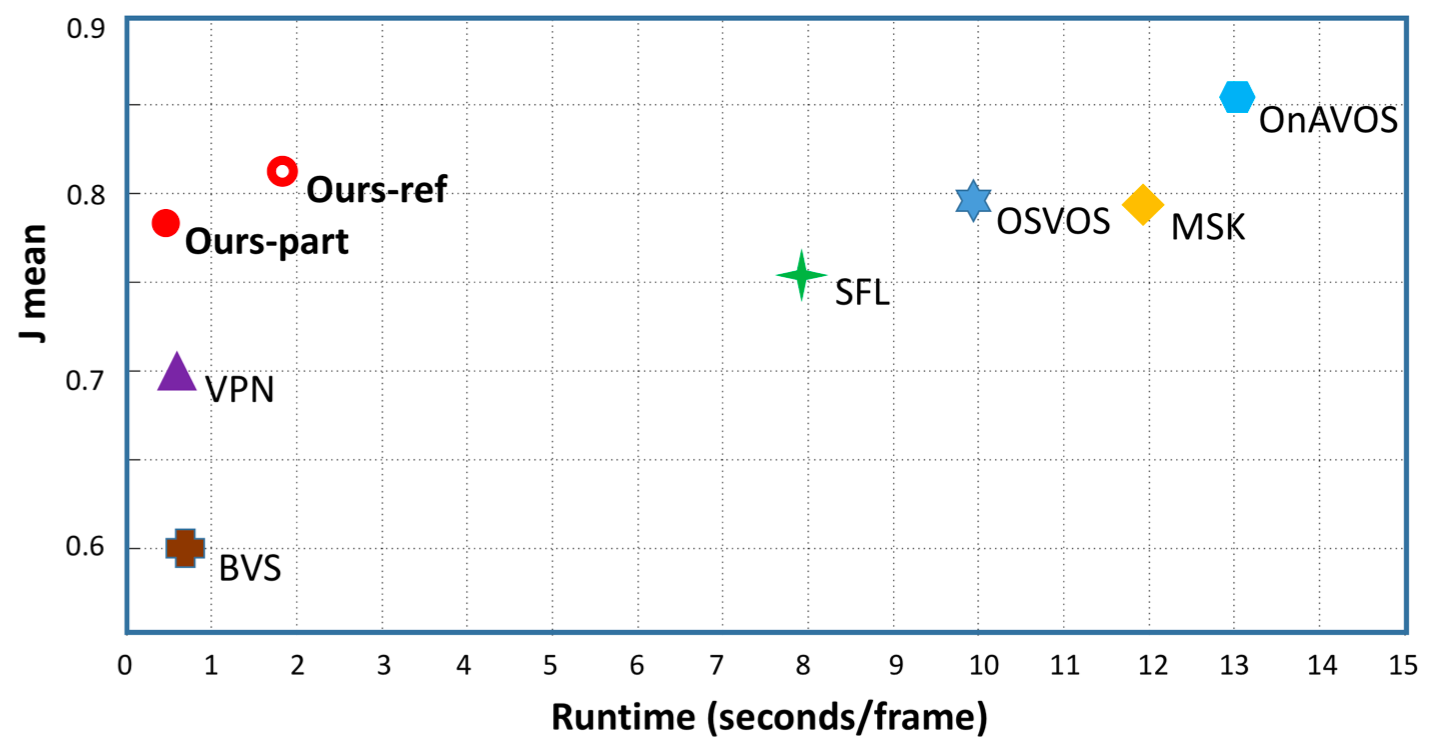}
		\end{center}
		\vspace{-3mm}
		\caption{ Accuracy versus runtime comparisons on the DAVIS 2016 dataset.
			We evaluate the state-of-the-art methods and demonstrate that our approach is significantly faster, while maintaining high accuracy.
			Note that the runtime includes the pre-processing steps averaged on all frames for fair comparisons.
		}
		\label{fig:runtime}
		 	\vspace{-6mm}
	\end{figure}

	Video object segmentation aims at separating target objects from the background and other instances on the pixel level.
	Segmenting objects in videos is a fundamental task in computer vision 
	because of its wide applications such as video surveillance, video editing, and autonomous driving.
	However, it is a challenging task due to camera motion, object deformation, occlusion between instances and cluttered background.
	Particularly for online applications, significant different issues arise when the methods are required to be robust and fast without given access to future frames.
	In this paper, we focus on solving the problem of online video object segmentation. 
	Given the object in the first frame, our goal is to immediately perform online segmentation on this target object without knowing future frames.
	For real application usages, the difficulties lie in the requirement of efficient runtime performance while maintaining accurate segmentation.
	Figure~\ref{fig:runtime} illustrates comparisons of the state-of-the-art methods in terms of speed and performance, where we show that the proposed algorithm is fast, accurate and applicable to online tasks.

	Existing video object segmentation algorithms can be broadly classified into unsupervised and semi-supervised settings.
	Unsupervised methods \cite{faktor2014video, jain2017fusionseg, keuper2015motion, Pap_ICCV_2013} mainly segment moving objects from the background without any prior knowledge of the target, e.g., initial object masks. 
	However, these methods cannot handle multiple object segmentation as they are not capable of identifying a specific instance.
	In addition, several methods require batch model processing (i.e., all the frames are available) before segmenting the object \cite{kohprimary, tokmakov2017learning}, which cannot be applied to online applications.
	\begin{figure*}[t]
		\begin{center}
			\includegraphics[width=0.95\linewidth]{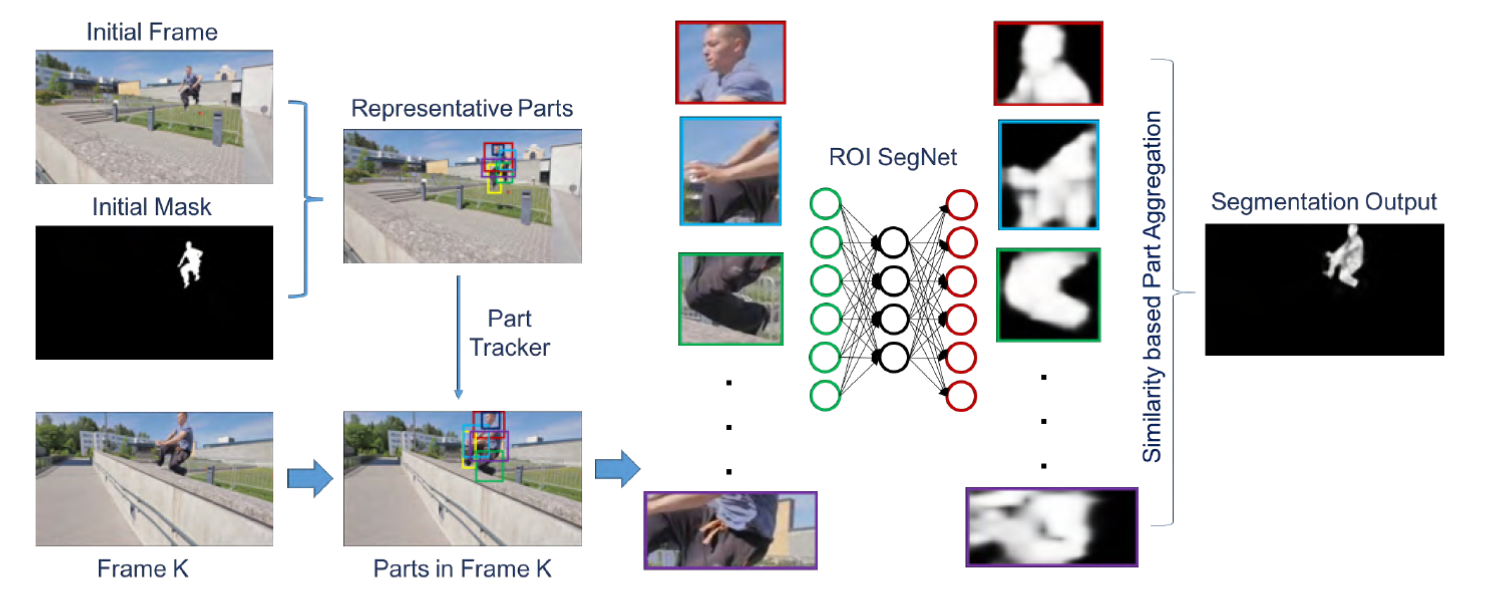}
		\end{center}
		\vspace{-3mm}
		\caption{Proposed framework for online video object segmentation. Our algorithm first generates parts of the target object in the first frame. These parts are then tracked in the next frame to obtain tracking boxes. With our ROI segmentation network and a similarity-based scoring function, final segmentation outputs are generated through the entire video.}
		\label{fig:framework}
		 \vspace{-4mm}
	\end{figure*}
	On the other hand, semi-supervised methods \cite{chengsegflow, jang2017online, khoreva2017lucid, khoreva2016learning, voigtlaender17BMVC} are given with an initial object mask which provides critical visual cues of the target.
	Thus, these methods can handle multi-instance cases and usually perform better than the unsupervised approaches. 
	However, many state-of-the-art semi-supervised methods heavily rely on the segmentation mask in the first frame. 
	For instance, before making predictions on the test video, the state-of-the-art methods need to finetune the networks for each video~\cite{caelles2016one, chengsegflow, khoreva2017lucid, voigtlaender17BMVC}, 
	or the model for each instance~\cite{cheng2017learning, DAVIS2017-8th}.
	This finetuning step on the video or instance level is 
	computationally expensive, where
	it usually takes more than ten minutes to update a model \cite{caelles2016one, chengsegflow}.
	In addition, data preparation (e.g., optical flow generation \cite{Tsai_CVPR_2016}) and training data augmentation \cite{khoreva2017lucid} require additional processing time.
	As such, these methods cannot be used for time-sensitive online applications that require fast and accurate segmentation results of a specific target object (see Figure \ref{fig:runtime}).

	In this paper, we propose a video object segmentation algorithm that can immediately start to segment a specific object through the entire video fast and accurately.
	To this end, we utilize a part-based tracking method and exploit a convolutional neural network (CNN) for representations but does not need the time-consuming finetuning stage on the target video.
	The proposed method mainly consists of three parts: part-based tracking, region-of-interest segmentation, and similarity-based aggregation.
	\vspace{-1mm}
	{\flushleft {\bf Part-based Tracking.}}
	Naturally, object tracking is an effective way to localize the target in the next frame.
	However, non-rigid objects often have large deformation with fast movement, thereby making it difficult to accurately localize the target \cite{bertinetto2016fully, danelljan2016eco, Ma_ICCV_2015}.
	To better utilize the tracking cues, we adopt a part-based tracking scheme to resolve challenging issues such as occlusions and appearance changes \cite{Liu_2015_CVPR}.
	We first randomly generate object proposals around the target in the first frame, and select representative parts based on the overlapping scores with the initial mask.
	We then apply the tracker for each part to provide temporally consistent region of interests (ROIs) for subsequent frames.
	\vspace{-1mm}
	{\flushleft {\bf ROI Segmentation.}}
	Once each part is localized in the next frame, we construct a CNN-based ROI SegNet to predict the segmentation mask that belongs to the target object.
	%
    Different from conventional foreground segmentation networks \cite{caelles2016one, chengsegflow, li2017fully} that focus on segmenting the entire object, our ROI SegNet learns to segment partial objects given the bounding box of part.
	\vspace{-1mm}
	{\flushleft {\bf Similarity-based Aggregation.}}
	With part tracking and ROI segmentation, the object location and segmentation mask can be roughly identified.
	However, there could be false positives due to incorrect tracking results.
	To reduce noisy segmentation parts, we design a similarity-based method to aggregate parts by computing the feature distance between tracked parts and the initial object mask.
	Figure \ref{fig:framework} shows the main steps of the proposed algorithm.

	To validate the proposed algorithm, we conduct extensive experiments with comparisons and ablation study on the DAVIS benchmark datasets \cite{Perazzi2016,Pont-Tuset_arXiv_2017}.
	We show that the proposed method performs favorably against state-of-the-art approaches in accuracy, while achieving much better runtime performance.
	The contributions of this work are as the following.
	First, we propose a fast and accurate video object segmentation method that is applicable to online tasks.
	Second, we develop the part-based tracking and similarity-based aggregation methods that effectively utilize the information contained in the first frame, without adding much computational load.
	Third, we design an ROI SegNet that takes bounding boxes of parts as the input, and outputs the segmentation mask for each part.

	\section{Related Work}
	\label{sec:related}
	
	{\flushleft {\bf Unsupervised Video Object Segmentation.}}
	Unsupervised video object segmentation methods aim to automatically discover and separate prominent objects from the background.
	These methods are based on probabilistic models~\cite{lee2011key, ma2012maximum}, motions~\cite{keuper2015motion, jang2016primary,  Pap_ICCV_2013}, and object proposals~\cite{Li_ICCV_2013, zhang2013video}.
	Existing approaches often rely on visual cues such as superpixels, saliency maps or optical flow to obtain initial object regions, and need to process the entire video in batch mode for refining object segmentation.
	In addition, generating and processing thousands of candidate regions in each frame is usually time-consuming.
	Recently, CNN-based methods~\cite{jain2017fusionseg, tokmakov2016learning, tokmakov2017learning} exploit learning rich hierarchical features (e.g., ImageNet pre-training) and large augmented data to achieve the state-of-the-art segmentation results. 
	However, these unsupervised methods are not able to segment a specific object due to motion confusions between different instances and dynamic background.
	
	{\flushleft {\bf Semi-supervised Video Object Segmentation.}} 
	Semi-supervised methods aim to segment a specific object with an initial mask.
	Numerous algorithms have been proposed based on tracking~\cite{Godec_ICCV_2011}, object proposals~\cite{perazzi2015fully}, graphical model~\cite{marki2016bilateral}, and optical flow~\cite{Tsai_CVPR_2016}.
	Similar to the unsupervised approaches, CNN-based methods~\cite{caelles2016one, chengsegflow, khoreva2016learning} have achieved significant improvement for video object segmentation.
	However, these methods usually heavily rely on finetuning models through the first frame~\cite{caelles2016one, khoreva2016learning}, data augmentation~\cite{khoreva2017lucid}, online model adaptation~\cite{voigtlaender17BMVC} and joint training with optical flow~\cite{chengsegflow}.
	These steps are computationally expensive (e.g., it takes more than 10 minutes for finetuning on the first frame in each video) and are not suitable for online vision applications.  
	
	To alleviate the issue of computational loads, a few methods are developed by propagating the object mask in the first frame through the entire video~\cite{jampani2016video, jang2017online}.
	Without exploiting much information in the first frame, these approaches suffer from the error accumulation after propagating a long period of time and thus do not perform as well as other methods.
	In contrast, the proposed algorithm incorporates part-based tracking and always keeps eyes on the first frame by a similarity-based part aggregation strategy.
	{\flushleft {\bf Object Tracking.}}
	Tracking has been widely used to localize objects in videos as an additional cue for performing object segmentation \cite{Tsai_ECCV_2016}.
	Conventional methods~\cite{bolme2010visual, henriques2015high} adopt correlation filter-based trackers to account for appearance changes.
	Recently, numerous methods have been developed based on deep neural networks and classifiers.
	The CF2 method~\cite{Ma_ICCV_2015} learns correlation filters adaptively based on CNN features, thereby enhancing the ability to handle challenging factors such as deformation and occlusion.
	In addition, the SINT scheme~\cite{tao2016siamese} utilizes a Siamese network to learn feature similarities between proposals and the initial observation of target object.
	The SiaFC algorithm~\cite{bertinetto2016fully} develops an end-to-end Siamese tracking network with fully-convolutional layers, which allows the tracker to compute similarity scores for all the proposals in one forward pass.
	In this work, we adopt the Siamese network for tracking object parts, where each part is locally representative and endures less deformation through the video.
	\section{Proposed Algorithm}
	In this section, we describe each component of the proposed method.
	First, we present the part-based tracker, where the goal is to localize object parts through the entire video.
	Second, we construct the ROI SegNet, a general and robust network to predict segmentation results for object parts.
	Third, we introduce our part aggregation method to generate final segmentation results by computing similarity scores in the feature space.
        \begin{figure*}[t]
        \vspace{-3mm}
		\begin{center}
			\includegraphics[width=1\linewidth]{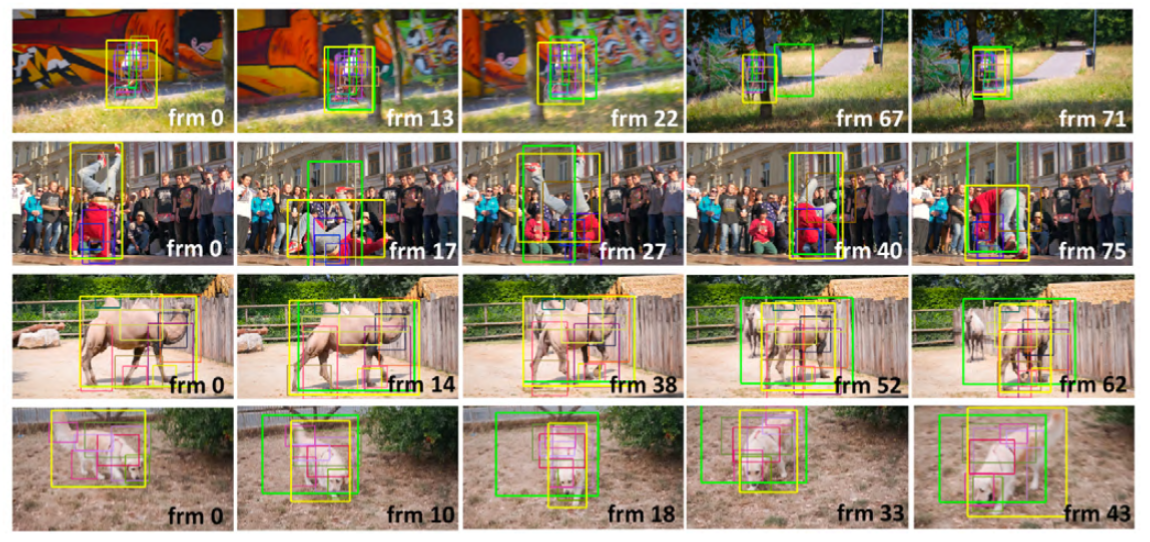}
		\end{center}
		\vspace{-3mm}
		\caption{Sample results for part tracking. 
			We show some high-scored parts and their tracking results.
			Green and yellow boxes are the results by applying object tracker \cite{bertinetto2016fully} and by our method via aggregating parts, respectively.
			It shows that our result (yellow boxes) are robust to object deformation and occlusion, due to the stability of tracking parts.
		}
		\label{fig:tracking}
        \vspace{-2mm}
	\end{figure*}
    \begin{figure}[t]
		\begin{center}
			\includegraphics[width=0.9\linewidth]{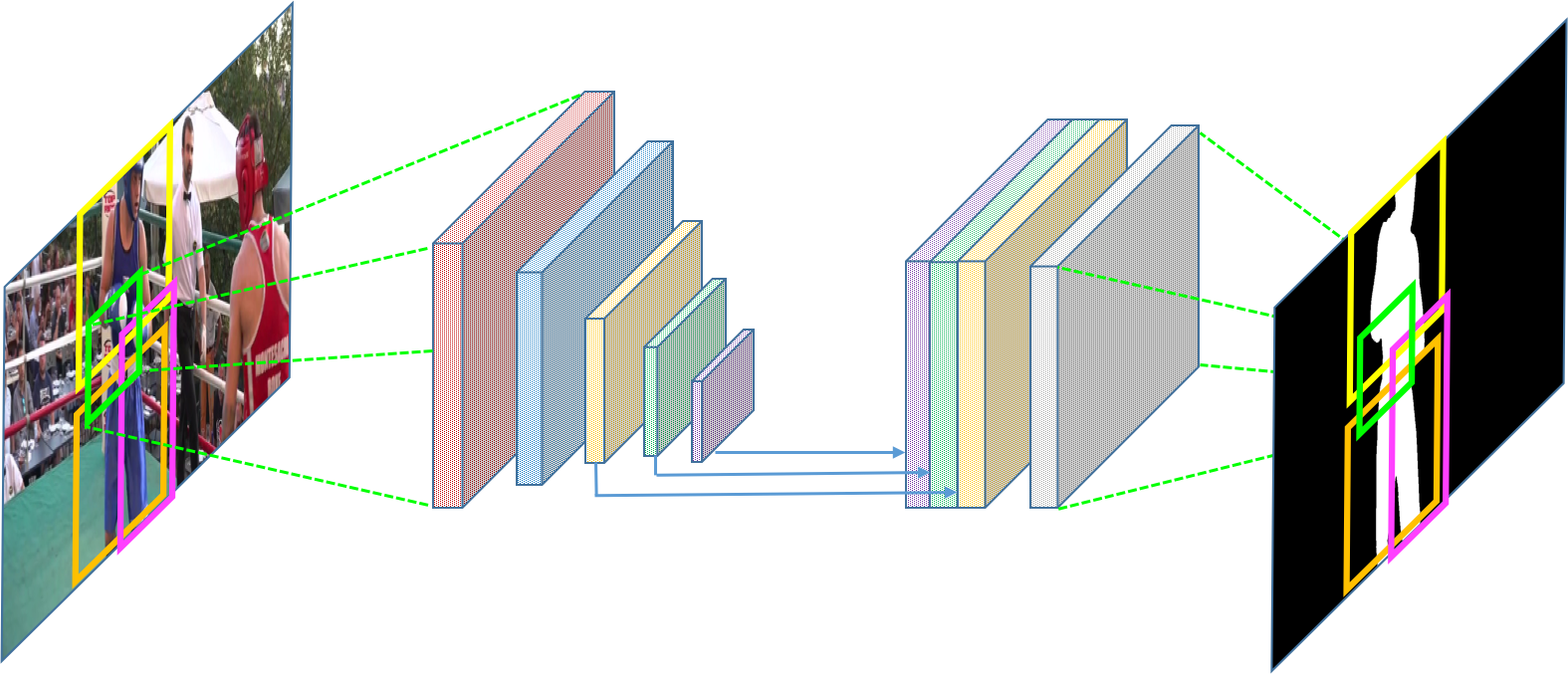}
		\end{center}
		\vspace{-3mm}
		\caption{Illustration of the proposed ROI SegNet.
	Given an image and their parts, we resize and align each part as the input to the network.
	We use the ResNet-101 architecture containing 5 convolution modules. We up-sample and concatenate feature maps from the last three modules. An additional convolution layer is utilized for the binary prediction of parts.
	}
		\label{fig:roi}
		\vspace{-3mm}
	\end{figure}
    
	\subsection{Part-based Tracking}
	Object tracking is a difficult task due to challenging factors such as object deformation, fast movement, occlusion, and background noise.
	To deal with these issues, part-based methods \cite{Liu_2015_CVPR} have been developed to track local regions instead of the entire object with larger appearance changes.
Since our goal is to localize most object regions in the next frame for further segmentation, utilizing a part-based method matches our need and can effectively maintain a high recall rate.
	{\flushleft {\bf Part Generation.}}
	In order to track parts, one critical problem is how to generate these parts in the first place.
Conventional object parts are discovered from a large amount of intra-class data via discriminability and consistency.
However, this assumption does not hold for online video segmentation, as only one object mask is provided in the first frame of the target video.
	To resolve this issue, we propose a simple yet effective way to generate representative parts guided by the object mask.
	First, we randomly generate part proposals with various sizes and locations around the object, and remove the ones with low overlapping ratio to the object mask.
	We compute the intersection-over-union (IoU) score between the proposal and the object, and keep the ones with scores larger than a threshold (i.e., $0.3$ in this work).
To ensure that each part contains mostly pixels from the object, we further measure the score: $\mathcal{S}_p =  \frac{bbox \cap gtbox}{bbox}$, where $bbox$ is the bounding box of a proposal and $gtbox$ is the known object box in the first frame.
	Part proposals with $\mathcal{S}_{p} > 0.7$ are used as candidates for a non-maximum suppression (NMS) step.
	Based on the proposed selection process, we reduce thousands of proposals to only $50\sim300$ representative parts depending on the object size.
	Note that, we also transform the bounding box for each part to be tight within the object mask, reducing background noise for more effective tracking and segmentation.
	Some example results are shown in Figure \ref{fig:tracking} for generated parts (with high scores) in the first frame.

	{\flushleft {\bf Part Tracking.}}
	Given a set of parts $\mathcal{P}_t = \{P_t^1, P_t^2,..., P_t^i \}$ in frame $I_t$, our goal is to output a score map $\mathcal{S}_t$ that measures the location likelihood of part $P_t^i$ appearing in the next frame $I_{t+1}$:
	\begin{equation}
	\mathcal{S}_t = \mathcal{T}(P_t^i, I_{t+1}),
	\end{equation}
	where $\mathcal{T}$ is a function to compute similarity scores between the part $P_t^i$ and the image $I_{t+1}$.
We use the SiaFC method~\cite{bertinetto2016fully} as our baseline tracker $\mathcal{T}$ to compute the score map $\mathcal{S}_t$.
	Due to its fully-convolutional architecture, we compute score maps for multiple parts in one forward pass.
	Once obtaining the score map, we select the bounding box with the largest response as the tracking result.
	Some tracking results are shown in Figure \ref{fig:tracking}.

	\subsection{ROI SegNet}
	\label{sec:ROISeg}
	Based on the tracking results of object parts, the next task is to segment partial object within the bounding box.
Recent instance-level segmentation methods \cite{he2017mask, dai2016instance} have demonstrated the state-of-the-art results by training networks for certain categories and output their segmentations. 
Our part segmentation problem is similar to the instance-level segmentation task but for the partial object. 
In addition, training such a network would require an alignment step for different parts as they may vary significantly in size, shape, and appearance for different instances or object categories.
	Hence, we utilize an ROI data layer by cropping image patches from parts as inputs to the network, in which these patches are aligned through resizing.
	Similar to semantic segmentation, our objective is to minimize the weighted cross-entropy loss for a binary (foreground/background) task:
	\begin{align}
	\mathcal{L}(P) = - (1-w)\sum_{i,j\in fg} \log \mathbb{E}(y_{ij} = 1;\theta) \notag \\
	- w\sum_{i,j\in bg} \log \mathbb{E}(y_{ij} = 0;\theta),
	\label{eq:seg}
	\end{align}
	\noindent
	where $\theta$ denotes CNN parameters, $y_{ij}$ denotes the network prediction for the input part $P$ at pixel $(i, j)$  and $w$ is the foreground-background pixel-number ratio used to balance the weights \cite{Xie_ICCV_2015}.

	{\flushleft {\bf Network Architecture.}}
	We utilize the ResNet-101 architecture \cite{he2016deep} as the base network for segmentation and transform it to fully-convolutional layers \cite{Long_CVPR_2015}.
	To enhance feature representations, we up-sample feature maps from the last three convolution modules and concatenate them together.
	The concatenated features are then followed by a convolution layer for the binary prediction.
	Figure \ref{fig:roi} shows the architecture of our ROI SegNet.
	
	{\flushleft {\bf Network Training.}}
	To train the proposed network, we first augment images from the training set of the DAVIS dataset \cite{Perazzi2016} via random scaling and affine transformations (i.e., flipping, $\pm 10\%$ shifting, $\pm 10\%$ scaling, $\pm 30^{\circ}$ rotation).
	Then, parts are extracted for each instance as the same method as introduced in part-based tracking.
	We use the Stochastic Gradient Descent (SGD) optimizer with the patch size $80 \times 80$ and the batch size of 100 for training.
	The initial learning rate starts from $10^{-6}$ and decreases by half for every 50,000 iterations. We train the network for 200,000 iterations.

	\subsection{Similarity-based Part Aggregation}
\begin{figure}[t]
		\begin{center}
			\includegraphics[width=1\linewidth]{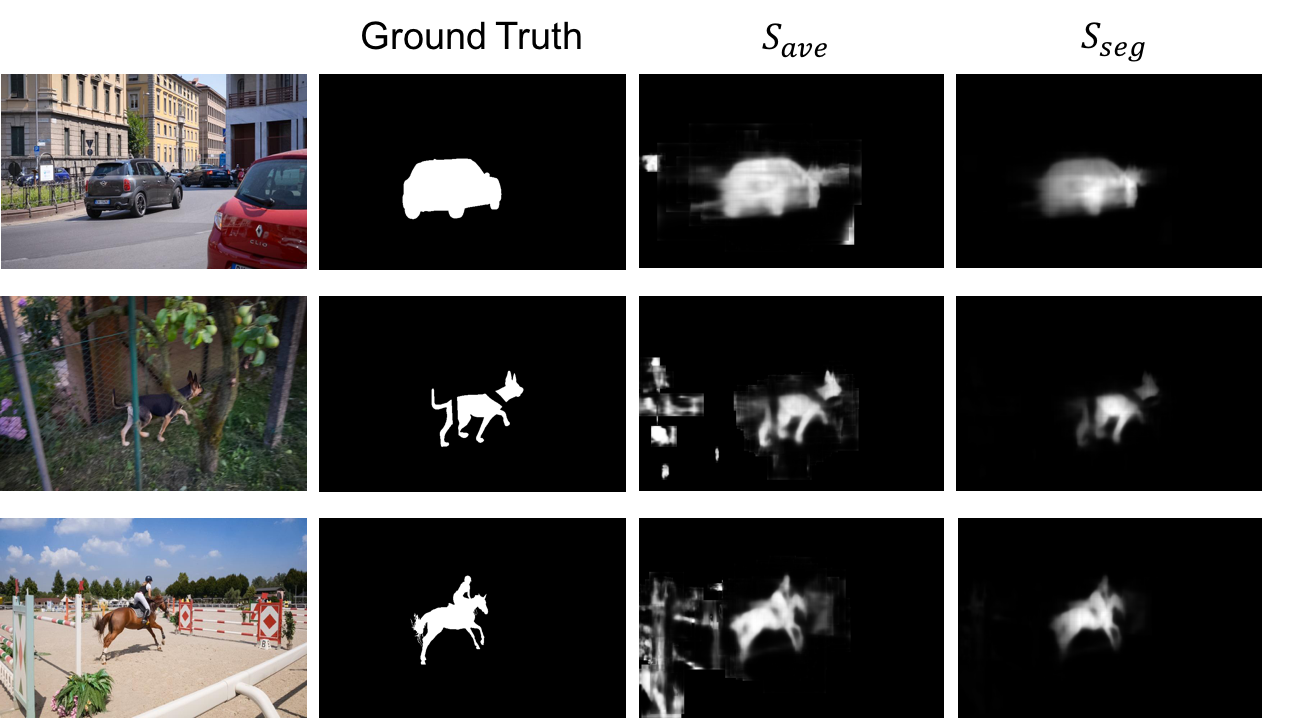}
		\end{center}
		\vspace{-3mm}
		\caption{Part aggregation results. 
			We compare score maps via the functions of $\mathcal{S}_{ave}$ and $\mathcal{S}_{seg}$.
			Without computing the similarity score to the first frame, the result of $\mathcal{S}_{ave}$ contains noisy segments,
while our aggregation algorithm performs segmentation more precisely.
		}
		\label{fig:ROIseg}
	\end{figure}

   After obtaining all the segmentation results from parts, one simple way to generate the final segmentation is to compute an averaging score map from each part.
	However, parts may be tracked off the object or include background noise, resulting in inaccurate part segments.
	To avoid adding these false positives, we develop a scoring function by looking back to the initial object mask.
	That is, we seek to know if the current part is similar to any of the parts in the first frame.
	Although objects may appear quite differently from the first frame, we find that local parts are actually more robust to such appearance changes.
	Specifically, we first compute the similarity score between each part in $\mathcal{P}_t$ at frame $t$ and initial parts $\mathcal{P}_0$ in the feature space.
	Then we select part $P^n_0$ with the highest similarity for the current part $P^m_t$ by:
	\begin{equation}
	n = \underset{i \in N}{\operatorname{argmin}} \| f(P^m_t)-f(P^i_0)) \|_2^2,
	\label{eq:part}
	\end{equation}
	where $f$ is the feature vector representing each part, extracted from the last layer in our ROI SegNet with an average pooling on the part mask.
	Overall, our scoring function consists of three components:
	\begin{equation}
	\mathcal{S}_{seg}(\mathcal{P}_t) = 	\mathcal{S}_{ave}(\mathcal{P}_t) \cdot \mathcal{S}_{sim}(\mathcal{P}_t, \mathcal{P}^n_0) \cdot \mathcal{S}_{con}(\mathcal{P}^n_0),
	\label{eq:part_score}
	\end{equation}
	where $\mathcal{P}^n_0$ is a set of initial parts selected based on Equation \eqref{eq:part} and $\cdot$ is the element-wise multiplication operation.
	The first function $\mathcal{S}_{ave}$ is the simple averaging score of part segments in the current frame $t$:
    %
    \begin{equation}
    \mathcal{S}_{ave}(\mathcal{P}_t) = \underset{i \in \mathcal{P}_t}\sum S^i / |\mathcal{P}_t|, 
	\label{eq:S-ave}
	\end{equation}
	where $\mathcal{P}_t$ is the set of parts at frame $t$ and $S^i$ is the segmentation score for each part $i$.
	Second, $\mathcal{S}_{sim}$ is the similarity score between current and initial parts in the feature space based on \eqref{eq:part}.
	Since the selected initial part segment may have poor quality, we add $\mathcal{S}_{con}$ by forwarding $\mathcal{P}^n_0$ to the ROI SegNet and measuring its segmentation overlapping ratio to the initial mask as the confidence score:
    \begin{equation}
	\mathcal{S}_{con}(\mathcal{P}^n_0) = J(G(\mathcal{P}^n_0), gt), 
	\label{eq:S-con}
	\end{equation}
	where $J$ is the IoU measurement, $G$ is the ROI SegNet and $gt$ is the object mask in the first frame.
	With the guidance of the initial object mask and parts without using expensive model finetuning step, our part aggregation method can effectively remove false positives.
	Figure \ref{fig:ROIseg} shows some examples of score maps with different scoring functions.

    \begin{figure}[t]
		\begin{center}
            \includegraphics[width=1\linewidth]{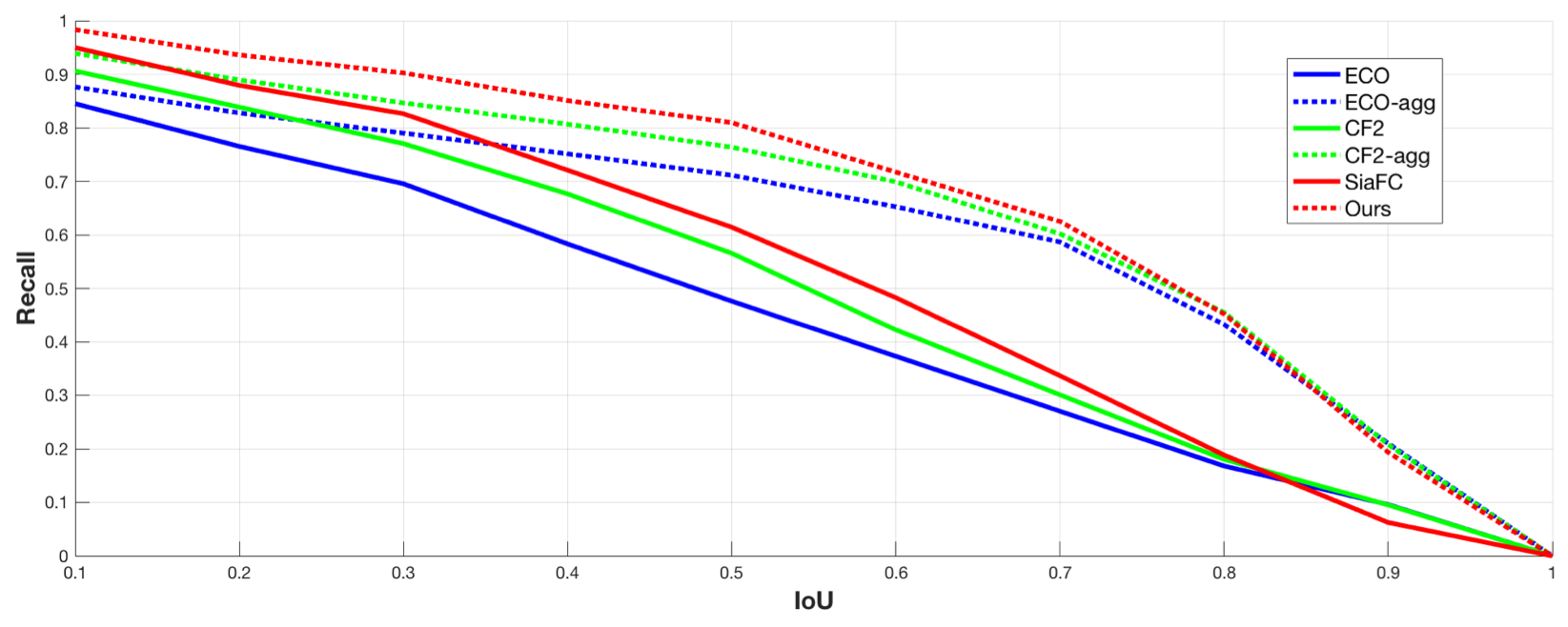}
		\end{center}
		\vspace{-3mm}
		\caption{IoU-Recall curve for trackers on the DAVIS 2016 dataset. Dashed lines (-agg) denote results by utilizing the proposed part-based tracking.}
		\label{fig:trackers}
		\vspace{-4mm}
	\end{figure}
	%

	\section{Experimental Results}
	\subsection{Dataset and Evaluation Metrics}
	We conduct experiments on the DAVIS benchmark datasets \cite{Pont-Tuset_arXiv_2017,Perazzi2016} which contain high-quality videos with dense pixel-level object segmentation annotations. 
	The DAVIS 2016 dataset consists of 50 sequences (30 for training and 20 for validation), with 3,455 annotated frames of real-world moving objects. 
Each video in the DAVIS 2016 dataset contains a single annotated foreground object, so both semi-supervised and unsupervised methods can be evaluated.
The DAVIS 2017 dataset contains 150 videos with 10,459 annotated frames and 376 object instances.
It is a challenging dataset as there are multiple instances in each video, where objects could occlude each other.
In this setting, it is difficult for unsupervised methods to separate different instances.
For performance evaluation, we use the mean region similarity ($J$ mean), contour accuracy ($F$ mean) and temporal stability ($T$ mean) as in the benchmark setting \cite{Pont-Tuset_arXiv_2017,Perazzi2016}.
The source code and models are available at \url{https://github.com/JingchunCheng/FAVOS}.
More results and analysis are presented in the supplementary material.

	\subsection{Tracker Evaluation}
	\label{sec:trackers}
	Our part-based tracker focuses on tracking local regions and cannot directly output the object location in the next frame.
	However, we can roughly find the object center based on the aggregated part segments.
	Motivated by the tracking-by-detection algorithms \cite{Babenko_CVPR_2009}, we utilize detection proposals \cite{liu2016ssd} as candidates of object bounding boxes, and select the one closest to the object center as the tracking result.
	We then validate this part-based tracker on the DAVIS 2016 dataset with comparisons to our baseline SiaFC method \cite{bertinetto2016fully} and other tracking algorithms including CF2 \cite{henriques2015high}, ECO \cite{danelljan2016eco}, and MDNet \cite{nam2016learning}.
Experimental results are presented in Figure \ref{fig:tracking} and \ref{fig:trackers}, where we show that our part-based trackers consistently maintain better IoU-recall curves for localizing objects.

	Although our ultimate goal is for video object segmentation, this evaluation is useful for understanding the challenges on the DAVIS dataset.
	One interesting fact is that if there is a good tracker, it should be able to help the segmentation task. 
	Thus, a high recall rate under a high IoU is required as once partial object is missing, it is not possible to recover the corresponding segment.
As shown in Figure \ref{fig:trackers}, most trackers achieve around 60\% recall rate under a 0.5 IoU while ours is 80\%, which enables potential usages of applying our tracker to improve segmentation results.
We will present our results by integrating this tracker in the ablation study section.
	\begin{table}[t]\scriptsize
		\caption{Ablation study on DAVIS 2016. 
        ``+ $\mathcal{S}_{seg}$'' and ``+ $\mathcal{S}_{seg}$ + Tracker + CRF'' denote results for {\bf Ours-part} and {\bf Ours-ref} in Figure \ref{fig:runtime}, respectively.
		}
				\vspace{-3mm}
		\vspace{1mm}
		\begin{center}
			\scriptsize
			\centering
			\renewcommand{\arraystretch}{1.5}
			\setlength{\tabcolsep}{4.5pt}
			\begin{tabular}{lcccccc}
				\toprule
				Method                & Baseline    & + \cite{bertinetto2016fully} &+ $\mathcal{S}_{ave}$&+ $\mathcal{S}_{seg}$&    + $\mathcal{S}_{seg}$&   + $\mathcal{S}_{seg}$         \\
				&          &&               &         &      + Tracker  &    + Tracker          \\
				&          &&               &         &             &   + CRF  \\
				\midrule
				J Mean $\uparrow$     &0.707     &0.696 & 0.739        &0.779    &0.786        &0.824\\
				J Recall $\uparrow$   &0.840     &0.860& 0.874         &0.924    &0.929        &0.965\\
				J Decay $\downarrow$  &-0.005    &0.008& 0.072         &0.067    &0.054        &0.045\\
				\midrule
				F Mean $\uparrow$     &0.695     &0.671& 0.727         &0.760    &0.772        &0.795\\
				F Recall $\uparrow$   &0.786     &0.790& 0.792         &0.849    &0.869        &0.894\\
				F Decay $\downarrow$  &-0.004    &-0.003& 0.089        &0.076    &0.060        &0.055\\
				\midrule
				T Mean $\downarrow$   &0.260     &0.321& 0.240         &0.229    &0.219        &0.263\\
				\bottomrule
			\end{tabular}
		\end{center}
		\label{tab:ablation}	
        \vspace{-5mm}
	\end{table}

	\begin{figure}[t]
		\begin{center}
			\includegraphics[width=1\linewidth]{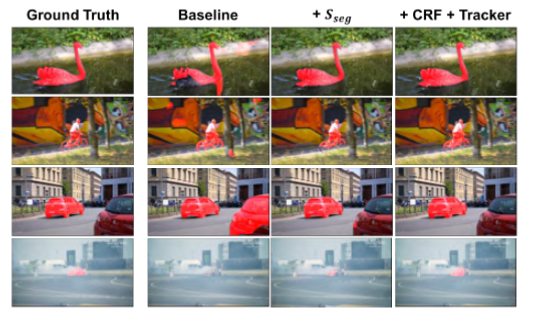}
		\end{center}
		\vspace{-3mm}
		\caption{Sample results of using different components in the proposed method. We show gradual improvement over baseline with part aggregation, CRF refinement and an object tracker.
		}
		\label{fig:ablation}
				\vspace{-4mm}
	\end{figure}
	\subsection{Ablation Study on Segmentation}
	\label{sec:ablation}
	\begin{table*}[t]\scriptsize
		\caption{Overall segmentation results on DAVIS 2016.
			We analyze various settings for different algorithms as well as provide online applicability based on their runtime speed (with different colors).
		}
		\vspace{1mm}
		\begin{center}	
			\small
			\centering
			\begin{tabular}{lcccccccc}
				\toprule
				
				Method      & Initial mask & Future frames  & Pre-processing & Online & Speed &J mean & F mean & T mean\\
				\midrule
				\rowcolor{mygray} OnAVOS \cite{voigtlaender17BMVC}    & \Checkmark &          & finetuning & weak   & 13s & 0.861 & 0.849 & 0.190\\
				\rowcolor{mygray} Lucid \cite{khoreva2017lucid}       & \Checkmark &          & data, finetuning  & weak & 40s & 0.848 & 0.823 & 0.158 \\
          \rowcolor{mypink} \bf Ours-ref                            & \Checkmark &          &  no   & strong & 1.8s     & 0.824 & 0.795 & 0.263\\
		  \rowcolor{mygray} OSVOS  \cite{caelles2016one}        & \Checkmark &          & finetuning & weak   & 10s & 0.798 & 0.806 & 0.378\\
		\rowcolor{mygray} MSK    \cite{khoreva2016learning}   & \Checkmark &          & flow, finetuning & weak   & 12s & 0.797 & 0.754 &0.218 \\    
          \rowcolor{mypink} \bf Ours-part                           & \Checkmark &          &  no   & strong & 0.60s     &0.779 & 0.760 & 0.229 \\
				ARP    \cite{kohprimary}            &           & \Checkmark & data  & no     &  -     & 0.762 & 0.706 &0.393\\
				\rowcolor{mygray} SFL    \cite{chengsegflow}          & \Checkmark &           & finetuning & weak   & 7.9s      & 0.761 & 0.760 &0.189\\
				LVO    \cite{tokmakov2017learning}  &           & \Checkmark & flow & no     &     -     & 0.759 & 0.721 &0.265\\
				\rowcolor{mygray} CTN    \cite{jang2017online}        & \Checkmark &           & flow & weak   & 29.95s    & 0.735 & 0.693 &0.220\\
				\rowcolor{mygray} FSEG  \cite{jain2017fusionseg}      &           &           & flow & weak   & 7s  & 0.707 & 0.653 & 0.328\\
				\rowcolor{mypink} VPN   \cite{jampani2016video}       & \Checkmark &           & no & strong & 0.63s     & 0.702 & 0.655 & 0.324\\
				\rowcolor{mygray} LMP   \cite{tokmakov2016learning}   &           &          & flow  & weak   & 18s & 0.700 & 0.659 & 0.572\\
				\rowcolor{mygray} OFL   \cite{Tsai_CVPR_2016}         & \Checkmark &           & flow & weak   & 60s & 0.680 & 0.634 & 0.222\\
				\rowcolor{mypink} BVS \cite{marki2016bilateral}       & \Checkmark &         & no & strong & 0.84s     & 0.600 & 0.588 & 0.347\\
				\bottomrule
			\end{tabular}
		\end{center}
		\label{tab:overall_davis}
		\vspace{-4mm}
	\end{table*}
    	\begin{figure*}[!th]
		\begin{center}
			\includegraphics[width=1\linewidth]{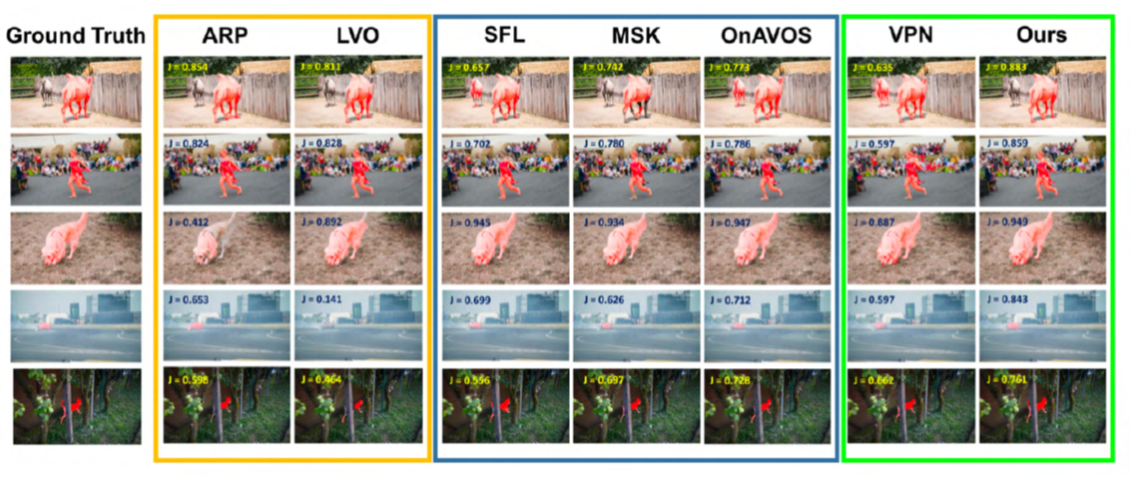}
		\end{center}
		\vspace{-3mm}
		\caption{
			Example results of comparisons between state-of-the-art methods on DAVIS 2016.
			Approaches with \textit{no}, \textit{weak}, \textit{strong} online applicability are marked in yellow, blue and green, respectively.}
		\label{fig:comparison}
		\vspace{-3mm}
	\end{figure*}

	We present ablation study in Table \ref{tab:ablation} on the DAVIS 2016 validation set to evaluate the effectiveness of each component in the proposed video object segmentation framework.
	%
    We start with the unsupervised version of SFL~\cite{chengsegflow} as our baseline due to its balance between speed and accuracy.
    To demonstrate the usefulness of using part, we first conduct an experiment by combining the baseline result and the score map from \cite{bertinetto2016fully} via tracking an entire object.
    Specifically, we average the foreground probability from \cite{chengsegflow} and the segmentation map of \cite{bertinetto2016fully} through the ROI SegNet.
	However, we find that the tracking accuracy is highly unstable, which usually loses objects and even results in a worse performance than the baseline segmentation (1.1\% drop in $J$ Mean).
	It shows that combining tracking and segmentation is not a trivial task, and we use part-based model to achieve a better combination.

	After adopting our part-based tracker and ROI SegNet to obtain part segments, we compare results with or without part aggregation.
	%
	%
	The one that utilizes part aggregation via the function $\mathcal{S}_{seg}$ in Equation \eqref{eq:part_score} performs better (4\% improvement in $J$ Mean) than only computing the score function $\mathcal{S}_{ave}$.
	It shows that with the consideration of initial object mask, false part segmentations can be largely reduced as they are not similar to any of the object parts in the first frame.
	In addition, we take advantage of our tracker combined with detection proposals as mentioned in Section \ref{sec:trackers} and use it to further improve our results, denoted as ``+Tracker'' in Table \ref{tab:ablation}.
    To further improve the boundary accuracy, we add a refinement step using dense CRF~\cite{KrahenbuhlK11}.
    In Figure \ref{fig:runtime}, we denote the result of using $\mathcal{S}_{seg}$ as \textit{Ours-part}, and the one combined with our tracker and CRF with refinement as \textit{Ours-ref}.

	\begin{table*}[t]\scriptsize
		\caption{Segmentation results on DAVIS 2017 validation set. 
			We show our baseline results with different modules, including foreground/background regularization (FG), Spatial Propagation Network (SPN) and a refinement procedure.
		}
		\vspace{-2mm}
		\begin{center}
			\small
			\centering
			\begin{tabular}{lccccccc}
				\toprule
				& Finetuning  & Method  &  Baseline  &+ FG   & + FG & +FG\\
				\multicolumn{2}{c}{}                &                    &       &    &  + SPN  &  +SPN \\
				\multicolumn{2}{c}{}                &          &           &          &         &   +Refine  \\
				\midrule
				Ours             &       & \multirow{2}{*}{J Mean $\uparrow$}       & 0.451    & 0.462      & 0.481   &0.546\\
				SPN \cite{cheng2017learning}&           \Checkmark          &  & 0.442     &0.457      &0.506    &0.540\\
				\midrule
				Ours              &      & \multirow{2}{*}{F Mean $\uparrow$}   & 0.554    & 0.571      & 0.574   &0.618\\
				SPN \cite{cheng2017learning}&           \Checkmark          &  &0.453     &0.504       &0.568    &0.611\\                              
				\bottomrule
			\end{tabular}
		\end{center}
		\label{tab:davis2017}
		\vspace{-5mm}
	\end{table*}
	\subsection{Segmentation Results}
	{\flushleft {\bf DAVIS 2016.}}
	We evaluate our proposed method on the validation set of DAVIS 2016~\cite{Perazzi2016} with comparisons to state-of-the-art algorithms, including semi-supervised and unsupervised settings.
	In Table \ref{tab:overall_davis}, we show results with different settings, including the need of initial object mask, future frames and pre-processing steps.
	Based on these requirements and their runtime speed, we then analyze the capability for online applications.

	For unsupervised methods that do not need the initial mask, they usually need to compute optical flow as the motion cue (FSEG \cite{jain2017fusionseg} and LMP \cite{tokmakov2016learning}) or foresee the entire video (LVO \cite{tokmakov2017learning} and ARP \cite{kohprimary}) to improve the performance, which is not applicable to online usages.
In addition, these methods cannot distinguish different instances and perform segmentation on a specific object.

In the semi-supervised setting, recent methods require various pre-processing steps before starting to segment the object in the video, which weaken the ability for online applications.
	These pre-processing steps include model finetuning (OnAVOS \cite{voigtlaender17BMVC}, Lucid \cite{khoreva2017lucid}, OSVOS \cite{caelles2016one}, MSK \cite{khoreva2016learning}, SFL \cite{chengsegflow}), data synthesis (Lucid \cite{khoreva2017lucid}) and flow computing (MSK \cite{khoreva2016learning}, CTN \cite{jang2017online}, and OFL \cite{Tsai_CVPR_2016}).
	For fair comparisons in the online setting, these pre-processing steps are included in the runtime by averaging on all the frames.

	The most closest setting to our method is VPN \cite{jampani2016video} and BVS \cite{marki2016bilateral} that do not have heavy pre-processing steps.
	However, these approaches may propagate segmentation errors after tracking for a long period of time. 
	In contrast, our algorithm always constantly refers to the initial object mask via parts and can reduce such errors in the long run, improving more than 12\% in $J$ Mean against VPN \cite{jampani2016video}.
	Overall, the proposed video object segmentation framework runs at the fastest speed, and can achieve $J$ Mean in the 3rd place with further refinement, while still maintaining a fast runtime speed compared to state-of-the-art methods.
	Some qualitative comparisons are presented in Figure \ref{fig:comparison}.
	{\flushleft {\bf DAVIS 2017.}}
	To evaluate how our method deals with multiple instances in videos, we conduct experiments on the DAVIS 2017 validation set~\cite{Pont-Tuset_arXiv_2017} which consists of 30 challenging videos and each one has two instances on average.
	%
	Existing methods all rely on sophisticated processing steps \cite{DAVIS2017-1st} to achieve better performance, and hence we compare our method with SPN \cite{cheng2017learning} that only involves the finetuning step in Table \ref{tab:davis2017}.
%
	For the baseline algorithm, we start with our part-based aggregation method via part-based tracker and ROI SegNet, while \cite{cheng2017learning} finetunes a CNN-based model for each instance.
	The baseline results show that, without the need of the computationally expensive finetuning process, our method even outperforms the existing method.
	One reason is that as the video becomes more complicated, finetuning-based methods may suffer from confusions between instances.
	In contrast, our method employs a part-based tracker that can effectively capture local cues for further segmentation.

\begin{figure}[t]
		\begin{center}
			\includegraphics[width=1\linewidth]{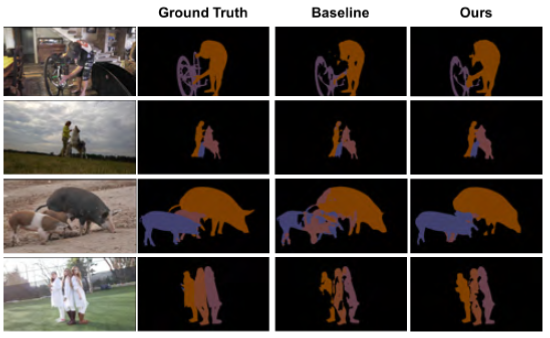}
		\end{center}
		\vspace{-3mm}
		\caption{Example results for multiple instances on DAVIS 2017.}
		\label{fig:multi}
		\vspace{-5mm}
	\end{figure}

	Following \cite{cheng2017learning}, we then sequentially add different components, including foreground/background regularization, a Spatial Propagation Network and a region-based refinement step.
	In addition, we integrate the object tracker proposed in Section \ref{sec:trackers} to further refine the segmentation.
	Overall, without the need of finetuning on each instance, our approach achieves a similar performance or outperforms the one that requires finetuning.
	We also note that finetuning is expensive not only in speed but also in stored size, as hundreds of objects would result in a huge number of stored models, which is not practical in real-world applications.
	In Figure \ref{fig:multi}, we present some example results on the DAVIS 2017 dataset.

	{\flushleft {\bf Runtime Analysis.}}
	In the proposed framework, our method runs at 0.60 seconds on average per instance per frame without the refinement step, including part-based tracking (0.2s), ROI segmentation (0.3s), and part aggregation (0.1s).
    %
    With CRF (1s) and tracker (0.2s) refinements, our method runs at 1.8 seconds per instance per frame with better performance.
	%
    We note that for tracking and segmenting parts, we parallelly use Titan X GPUs to handle hundreds of parts for faster inference.

	%
	\section{Concluding Remarks}
	In this paper, we propose a fast and accurate video object segmentation method that is applicable to online applications.
	Different from existing algorithms that heavily rely on pre-processing the object mask in the first frame, our method exploits the initial mask via a part-based tracker and an effective part aggregation strategy.
	The part-based tracker provides good localization for local regions surrounding the object, ensuring that most portion of the object is retained for further segmentation purpose.
We then design an ROI segmentation network to accurately output partial object segmentations.
Finally, a similarity-based scoring function is developed to aggregate parts and generate the final result.
	Our algorithm exploits the strength of CNN-based frameworks for tracking and segmentation to achieve fast runtime speed, while closely monitoring the information contained in the first frame for the state-of-the-art performance.
The proposed algorithm can be applied to other video analytic tasks that require fast and accurate online video object segmentation. 
%

\vspace{-3mm}
{\flushleft {\bf Acknowledgments.}}
This project is supported in part by the NSF CAREER Grant \#1149783, gifts from Adobe and NVIDIA.
	
	{\small
		\bibliographystyle{ieee}
		\bibliography{mybib}
	}
	
\end{document}